\documentclass{article}

\usepackage[nonatbib, preprint]{neurips_2021}

\usepackage[utf8]{inputenc} 
\usepackage[T1]{fontenc}    
\usepackage{url}            
\usepackage{booktabs}       
\usepackage{amsfonts}       
\usepackage{nicefrac}       
\usepackage{microtype}      
\usepackage{xcolor}         
\usepackage{amsmath}
\usepackage{xspace}
\usepackage{epsfig}
\usepackage{graphicx}
\usepackage{caption}
\usepackage{subcaption}

\def\mI{{\mathcal I}}

\def\0{{\bf 0}}
\def\1{{\bf 1}}

\def\bW{{\bf W}}
\def\bX{{\bf X}}

\usepackage{multirow}
\usepackage{array}
\usepackage[switch]{lineno}

\newcommand{\eg}{\textit{e}.\textit{g}.}
\newcommand{\etal}{\textit{et al}. }

\definecolor{citecolor}{HTML}{0071bc}
\usepackage[pagebackref=false,breaklinks=true,colorlinks,citecolor=citecolor,bookmarks=false]{hyperref}

\newlength\savewidth\newcommand\shline{\noalign{\global\savewidth\arrayrulewidth
  \global\arrayrulewidth 1pt}\hline\noalign{\global\arrayrulewidth\savewidth}}
  
  \makeatletter
\def\@fnsymbol#1{\ensuremath{\ifcase#1\or \dagger\or \ddagger\or
   \mathsection\or \mathparagraph\or \|\or **\or \dagger\dagger
   \or \ddagger\ddagger \else\@ctrerr\fi}}
\makeatother

\definecolor{mypink}{rgb}{0.858, 0.188, 0.478}

\pdfoutput=1
  
\title{Less is More: Pay Less Attention in Vision Transformers}

\author{Zizheng Pan, \quad Bohan Zhuang\thanks{Corresponding author. E-mail: $\tt  bohan.zhuang@monash.edu$}, \quad Haoyu He,  \quad Jing Liu, \quad Jianfei Cai \\[0.2cm]
Data Science \& AI, Monash University, Australia
}

\begin{document}

\maketitle

\begin{abstract}

Transformers \cite{transformer} have become one of the dominant architectures in deep learning, particularly as a powerful alternative to convolutional neural networks (CNNs) in computer vision. However, Transformer training and inference in previous works can be prohibitively expensive due to the quadratic complexity of self-attention over a long sequence of representations, especially for high-resolution dense prediction tasks. 
To this end, we present a novel Less attention vIsion Transformer (LIT), 
building upon the fact that the early self-attention layers in Transformers still focus on local patterns and bring minor benefits in recent hierarchical vision Transformers.
Specifically, we propose a hierarchical Transformer where we use pure multi-layer perceptrons (MLPs) to encode rich local patterns in the early stages while applying self-attention modules to capture longer dependencies in deeper layers. Moreover, we further propose a learned deformable token merging module to adaptively fuse informative patches in a non-uniform manner.
The proposed LIT achieves promising performance on image recognition tasks, including image classification, object detection and instance segmentation, serving as a strong backbone for many vision tasks.  
Code is available at:
\def\UrlFont{\sf}
    \def\UrlFont{\rm\small\ttfamily}
\url{https://github.com/zhuang-group/LIT}

\end{abstract}

\section{Introduction}

Transformers have made substantial strides in natural language processing (NLP) (\eg, \cite{transformer, devlin2019bert}) and recently in the computer vision (CV) field (\eg, \cite{vit, deit}). Inspired by the pyramid design~\cite{fpn} in CNNs, recent hierarchical vision Transformers (HVTs)~\cite{pvt,swin,cvt,contnet} divide transformer blocks into several stages and progressively shrink feature maps as the network goes deeper. However, high-resolution feature maps in the early stages result in long token sequences, which brings huge computational cost and memory consumption due to the quadratic complexity of self-attention. For instance, a feature map of size $56 \times 56 \times 96$ costs 2.0G FLOPs in one Multi-head Self-Attention (MSA)~\cite{transformer}, while the entire model of ResNet-18~\cite{resnet} only requires 1.8G FLOPs. Such a huge computing cost makes it difficult to apply Transformers into broad computer vision tasks. 

Several emerging efforts have been made to reduce the computational cost in the early stages of HVTs. For example, some works~\cite{cvt, halonet} reduce the number of self-attention heads in an MSA layer or further decreasing the number of Transformer blocks~\cite{vit}. Another line of works proposes to trade-off accuracy and efficiency for MSA via heuristic approximation, such as spatial reduction attention (SRA)~\cite{pvt} and shifted window based multi-head self-attention (SW-MSA)~\cite{swin}. There are also studies simply employ convolutional layers~\cite{botnet, levit} when the resolution of feature maps are considerably large. However, how much the early adoption of self-attention layers really contributes to the final performance remains unclear.

In this paper, we present a \textbf{L}ess attention v\textbf{I}sion \textbf{T}ransformer (\textbf{LIT}) to address the aforementioned problem for HVTs. Specifically, we propose to exclusively use MLP layers to capture local patterns in the early stages while introducing MSA layers with sufficient number of heads to handle long range dependencies in the later stages. 
The motivation comes from two aspects. 
First, previous studies in both CNNs and Transformers have shown that shallow layers focus on local patterns and deeper layers tend to 
capture high-level semantics or global relationships~\cite{WuSH19,HouCHBTT19,local_bert}, arising the question of whether using self-attention at the early stages is necessary.
Second, from the theoretical perspective, a self-attention layer with sufficient heads applied to images can express any convolutional layer~\cite{selfconv}. 
However, fewer heads in an MSA layer theoretically hinder the ability of approximating a convolutional layer with a large kernel size,
where the extreme case is as expressive as a $1 \times 1$ convolution that can be viewed as a standard FC layer applied to each pixel independently. While recent HVTs adopt very few heads at the early stages to deliver pyramid representations, we argue that this is not optimal as such setting introduces high computational and memory cost but brings minor benefits.

To be emphasized, by exploiting MLP blocks in the early stages, the model avoids the huge computational cost and memory footprint arising from self-attention on high-resolution feature maps. Moreover, applying self-attention in the later stages to capture long range dependencies is quite efficient due to the progressive shrinking pyramid. Our comprehensive results show that such a simple architecture design brings a sweet spot between model performance and efficiency.

Furthermore, recent HVTs either adopt a standard convolutional layer or a linear projection layer to merge nearby tokens~\cite{pvt, swin}, aiming to control the scale of feature maps. However, such methods hinder the representational power for vision Transformers to model geometric transformations, considering that not every pixel equally contributes to an output unit~\cite{LuoLUZ16}. To this end, we propose a Deformable Token Merging (DTM) module, inspired by deformable convolutions~\cite{dconv1,dconv2}, where we learn a grid of offsets to adaptively augment the spatial sampling locations for merging neighboring patches from a sub-window in a feature map. In this way, we can obtain more informative downsampled tokens for subsequent processing.

Our contributions can be summarized as follows. First, we 
identify the minor contribution of the early MSA layers in recent HVTs and 
propose a simple HVT structure with pure MLP blocks in the early stages.
Second, we propose a deformable token merging module to adaptively merge more informative patches to deliver hierarchical representations, with enhanced transformation modeling capability. Finally, we conduct extensive experiments to show that the proposed LIT performs favorably against 
several state-of-the-art vision Transformers 
with similar or even reduced computational complexity and memory consumption.

\section{Related Work}

\paragraph{Vision Transformers.} 
Vision Transformers are models which adopt the self-attention mechanism ~\cite{transformer} into CV tasks. Recent works towards Vision Transformers either follow a hybrid architecture that combines convolution and self-attention~\cite{detr,botnet}, or design a pure self-attention architecture without convolution~\cite{sasa,lrnet}.
More recently, Dosovitskiy~\etal~\cite{vit} propose a Vision Transformer (ViT) which achieves promising results on ImageNet. 
Since then, a few subsequent works have been proposed to improve ViT from different aspects. 
For example, some works~\cite{t2t,localvit} seek to bring locality into ViT as they find ViT failed to model the important local structures (\eg, edges, lines). 
Another line of works aims to explore deeper architectures~\cite{deepvit, godeeper} by stacking more Transformer blocks. Some studies~\cite{glit,AutoFormer} also try to search a well-performed ViT with neural architecture search (NAS).

There is also a prevailing trend to introduce hierarchical representations into ViT~\cite{hvt,ceit,pvt,swin,cvt,pit,halonet}. 
To do so, these works divide Transformer blocks into several stages and downsample feature maps as the network goes deeper.
However, high-resolution feature maps in the early stages inevitably result in high computational and memory costs due to the quadratic complexity of the self-attention module. 
Targeting at this problem, Wang~\etal~\cite{pvt} propose to reduce the spatial
dimensions of attention's key and value matrices. Liu~\etal~\cite{swin} propose to limit self-attention in non-overlapped local windows. However, these replacements seek to shrink the global attention maps for efficiency. 
In this paper, we elaborately design the shallow layers with pure MLPs, that are powerful enough to encode local patterns. 
This neat architecture design keeps the capability for modelling global dependencies in the later stages while easing the prohibitively expensive complexity introduced by high-resolution feature maps, especially in the dense prediction tasks.

\paragraph{Deformable Convolutions.} 
Deformable convolutions (DC) are initially proposed by Dai~\etal~\cite{dconv1} in  object detection and semantic segmentation tasks. Different from the regular sampling grid of a standard convolution, DC adaptively augments the spatial sampling locations with learned offsets. One following work by Zhu~\etal~\cite{dconv2} improves DC with a modulation mechanism, which modulates the input feature amplitudes from different spatial locations. 
With the advantage on modeling geometric transformations, many works adopt DC to target various CV problems. For example,
Zhu~\etal~\cite{deformable_detr} propose a deformable attention module for object detection. Shim~\etal~\cite{refsr} construct a similarity search and extraction network built upon DC layers for single image super-resolution. Thomas~\etal~\cite{kpconv} introduce a deformable kernel point convolution operator for point clouds.
In this paper, we propose a deformable token merging module to adaptively merge more informative image tokens. Unlike previous works that merge tokens from a regular grid, DTM introduces better transformation modeling capability for HVTs.

\begin{figure*}[htp!]
	\centering
	\includegraphics[width=0.9\linewidth]{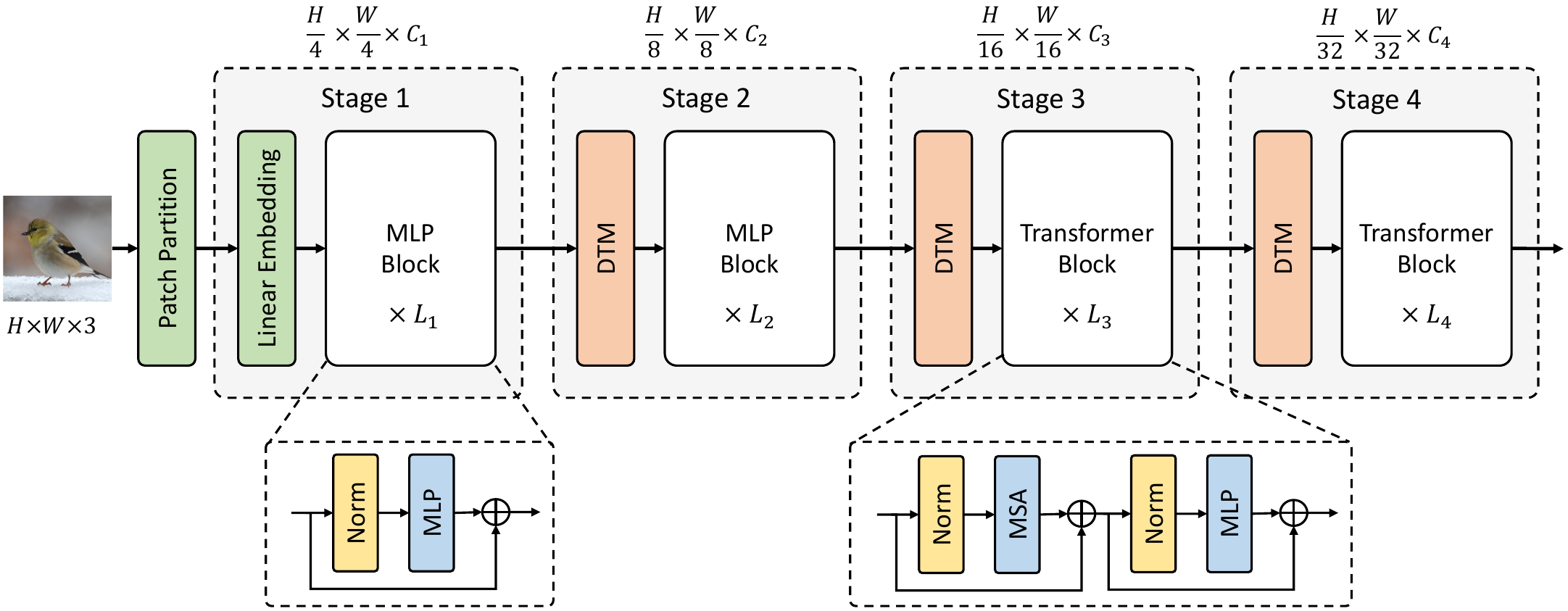}
	\caption{\textbf{Overall architecture of LIT.} The model is divided into four stages, where we apply MLP blocks in the first two stages and employ standard Transformer blocks in the last two stages. ``DTM'' denotes our proposed deformable token merging module.}
	\label{fig:model_arch}
\end{figure*}
\section{Proposed Method} \label{method}

\subsection{Overall Architecture}
\label{overall}
The overall architecture of LIT is illustrated in Figure~\ref{fig:model_arch}. Let $\mI \in \mathbb{R}^{H \times W \times 3}$ be an input RGB image, where $H$ and $W$ represent the height and width, respectively. We first split $\mI$ into non-overlapping patches with a patch size of $4\times4$, and thus the initial feature dimension of each patch is $4 \times 4 \times 3 = 48$. Next, a linear embedding layer is exploited to project each patch into dimension $C_1$, serving as the initial input for the following pipeline. The entire model is divided into 4 stages. Letting $s\in [1,2,3,4]$ be the index of a stage, we employ $L_s$ blocks at each stage, where the first two stages solely utilise MLP blocks to encode local representations and the last two stages employ standard Transformer blocks~\cite{vit} to handle longer dependencies. At each stage, we scale the input feature maps into $\frac{H_{s-1}}{P_s} \times \frac{W_{s-1}}{P_s} \times C_s$, where $P_s$ and $C_s$ represent the patch size and the hidden dimension at the $s$-th stage, respectively. For the last two stages, we set $N_s$ self-attention heads in each Transformer block.

\subsection{Block Design in LIT} \label{blockdesign}
As shown in Figure~\ref{fig:model_arch}, LIT employs two types of blocks: MLP blocks and Transformer blocks.
In the early stages, we apply MLP blocks. Concretely, an MLP block is built upon an MLP which consists of two FC layers with GELU~\cite{gelu} non-linearity in between. For each MLP at the $s$-th stage, an expansion ratio of $E_s$ is used. Specifically, the first FC layer expands the dimension of a token from $C_s$ to $E_s \times C_s$, and the other FC layer reduces the dimension back to $C_s$. Formally, letting $\bX \in \mathbb{R}^{(\frac{H_{s-1}}{P_s} \times \frac{W_{s-1}}{P_s}) \times C_s}$ be the input of the $s$-th stage and $l$ be the index of a block, an MLP block can be formulated as
\begin{linenomath}
\begin{equation}
    \bX_l = \bX_{l-1} + \mathrm{MLP}(\mathrm{LN}(\bX_{l-1})),
\end{equation}
\end{linenomath}
where $\mathrm{LN}$ indicates the layer normalization~\cite{layernorm} and $\mathrm{MLP}$ denotes an MLP.
In the last stages, a Transformer block as described in ViT~\cite{vit} contains an MSA layer and an MLP, which can be expressed as
\begin{linenomath}
\begin{align}
    \label{eq:msa}
    \bX^{'}_{l-1} &= \bX_{l-1} + \mathrm{MSA}(\mathrm{LN}(\bX_{l-1})), \\
    \label{eq:mlp}
    \bX_l &= \bX^{'}_{l-1} + \mathrm{MLP}(\mathrm{LN}(\bX^{'}_{l-1})).
\end{align}
\end{linenomath}
With this architecture, our model benefits from two main advantages: First, we avoid the
huge computational costs and memory footprint that are introduced by long sequences
in the early stages. Second, unlike recent works that shrink the attention maps using sub-windows~\cite{swin} or reduce the spatial dimensions of the key and value matrices, we keep standard MSA layers in the last two stages so as to maintain the capability of LIT to handle long range dependencies while keeping mild FLOPs due to the pyramid structure. We will show in the ablation study that our simple architecture design outperforms the state-of-the-art hierarchical ViT variants on ImageNet with comparable FLOPs.

\paragraph{Remark.}
Here we justify the rationality of applying pure MLP blocks in the early stages by considering the relationship among a convolutional layer, an FC layer and an MSA layer.
Firstly, we begin with a review of a standard convolutional layer. Let $\bX \in \mathbb{R}^{H\times W \times C_{in}}$ be the input feature map, and let $\bW \in \mathbb{R}^{K\times K \times C_{in} \times C_{out} }$ be the convolutional weight tensor, where $K$ is the kernel size, $C_{in}$ and $C_{out}$ are the input and output channel dimensions, respectively. For simplicity, we omit the bias term and use $\bX_{p,:}$ to represent  $\bX_{i,j,:,:}$, where $(i,j)$ denotes the pixel index and $p \in [H]\times[W]$. Given a convolutional kernel of $K\times K$ sampling locations, the output for a pixel $p$ can be formulated as
\begin{linenomath}
\begin{equation} \label{conv}
    \mathrm{Conv}(\bX)_{p,:} = \sum_{k \in [K\times K]} \bX_{p + g(k),:}\bW_{g(k),:,:},
\end{equation}
\end{linenomath}
where $g : [K \times K]\rightarrow\Delta_K$ is a bijective mapping of sampling indexes onto the pre-specified offsets $\Delta_K$. For example, let $\Delta_K = \{(-1,-1), (-1,0), ..., (0,1), (1,1)\}$ be a $3 \times 3$ kernel with dilation 1, then $g(0) = (-1, -1)$ represents the first sampling offset.

When $K = 1$, the weight tensor $\bW$ is equivalent to a matrix, such that $\bW \in \mathbb{R}^{C_{in}\times C_{out}}$. In this case, Eq.~(\ref{conv}) can express an FC layer, and the output for a pixel $p$ is defined by
\begin{linenomath}
\begin{equation} \label{fc}
    \mathrm{FC}(\bX)_{p,:}  = \bX_{p,:}\bW_{:,:}.
\end{equation}
\end{linenomath}
Last, let $N_h$ be the number of heads in an MSA layer and $\bW^{(h)} \in \mathbb{R}^{C_{in}\times C_{out}}$ be learnable parameters of the $h$-th head. 
Under a specific relative positional encoding scheme, 
Cordonnier~\etal~\cite{selfconv} prove that the output from an MSA layer at pixel $p$ 
can be formulated as 
\begin{linenomath}
\begin{equation} \label{re_msa}
    \mathrm{MSA}(\bX)_p = \sum_{h \in [N_h]}\bX_{p+f(h),:} \bW^{(h)},
\end{equation}
\end{linenomath}
where $f : [N_h]  \rightarrow \Delta_K$ is a bijective mapping of heads onto pixel shifts. In that case,
Eq.~(\ref{re_msa}) can be seen 
as an approximation to
a convolutional layer with a kernel size of $\sqrt{N_h}\times\sqrt{N_h}$. We refer detailed explanations to \cite{selfconv}.

From Eqs.~(\ref{conv})-(\ref{re_msa}), we observe that while an MSA layer with sufficient heads is able to approximate any convolutional layer, fewer heads theoretically limit the ability of such approximation. As an extreme case, an MSA layer with one head is only capable of approximating an FC layer.
Note that an MSA layer is certainly not equivalent to a convolutional layer in practice. However, d’Ascoli~\etal~\cite{convit} observe that the early MSA layers can learn to behave convolutional upon training. 
Considering most recent HVTs~\cite{pvt,cvt} adopt very few heads in the early stages, such convolutional behavior could be limited within small receptive fields.
In Figure~\ref{fig:attn_prob_vis}, we show in visualizations that the early MSA layers in PVT-S~\cite{pvt} indeed only attend to a tiny area around the query pixel, while removing them costs a minor performance drop but achieves significant reduction in model complexity. This justifies our method of applying pure MLP blocks in the first two stages without self-attention.

\subsection{Deformable Token Merging}
Previous works on HVTs~\cite{pvt,swin} rely on patch merging to achieve pyramid feature representations. However, they merge patches from a regular grid and neglect the fact that not every patch contributes equally to an output unit~\cite{LuoLUZ16}. Inspired by deformable convolutions~\cite{dconv1,dconv2}, we propose a deformable token merging module to learn a grid of offsets to adaptively sample more informative patches. 
Formally, a deformable convolution is formulated as
\begin{linenomath}
\begin{equation} \label{dcn}
    \mathrm{DC}(\bX)_{p,:} = \sum_{k \in [K\times K]} \bX_{p + g(k) + \Delta g(k),:}\bW_{g(k),:,:}.
\end{equation}
\end{linenomath}
Compared to a standard convolution operation as in Eq.~(\ref{conv}), DC learns an offset $\Delta g(k)$ for each pre-specified offset $g(k)$. Learning $\Delta g(k)$ requires a separate convolutional layer, which is also applied over the input feature map $\bX$. To merge patches in an adaptive manner, we adopt one DC layer in a DTM module, which can be formulated by 
\begin{linenomath}
\begin{equation}
    \mathrm{DTM}(\bX) =  \mathrm{GELU}( \mathrm{BN}( \mathrm{DC}(\bX))),
\end{equation}
\end{linenomath}
where $\mathrm{BN}$ denotes the batch normalization~\cite{batchnorm} and we employ the GELU non-linearity.
We will show in the ablation study that the sampling locations in DTM are adaptively adjusted when objects' scales and shapes change, benefiting from the learned offsets.
Also note that our light-weight DTM introduces negligible FLOPs and parameters compared to regular grid sampling in baselines, thus making it a plug-and-play module for recent HVTs.

\section{Experiments} \label{experiments}

\subsection{ImageNet Classification}
We conduct experiments on ImageNet (ILSVRC2012) \cite{imagenet} dataset. ImageNet is a large-scale dataset which has $\sim$1.2M training images from 1K categories and 50K validation images. We compare with CNN-based ResNet \cite{resnet} and Transformer-based models including DeiT~\cite{deit}, PVT~\cite{pvt} and Swin~\cite{swin}. For simplicity, we denote them as ``Model-Ti/S/M/B'' to refer to their tiny, small, medium and base variants. Similarly, we define four variants of our LIT models: LIT-Ti, LIT-S, LIT-M and LIT-B. 
Detailed architecture specifications are included in the supplementary material.
For better comparison, we design LIT-Ti as a counterpart to PVT-S, where both models adopt the absolute positional encoding. Our LIT-S, LIT-M and LIT-B use the relative positional encoding, and these three models can be seen as competitors to Swin-Ti, Swin-S, Swin-B, respectively. 

\paragraph{Implementation details.}
In general, all models are trained on ImageNet with 300 epochs and a total batch size of 1024. For all ImageNet experiments, training images are resized to $256 \times 256$, and $224 \times 224$ patches are randomly cropped from an image or its horizontal flip, with the per-pixel mean subtracted. We use the single-crop setting for testing. We use AdamW optimizer~\cite{adamw} with a cosine decay learning rate scheduler. The initial learning rate is $1\times10^{-3}$, and the weight decay is set to $5\times10^{-2}$. The initial values of learnable offsets in DTM are set to 0, and the initial learning rate for offset parameters is set to $1\times10^{-5}$. The kernel sizes and strides in DTM are consistent with that of patch merging layers in PVT and Swin.
For a fair comparison, we adopt the same training strategies as PVT and Swin when comparing our models to each of them. 

\begin{table*}[]
\centering
\renewcommand\arraystretch{1.2}
\caption{Comparisons with several state-of-the-art methods on ImageNet~\cite{imagenet}, ImageNet-Real~\cite{imagenet_real} and ImageNet-V2 matched frequency~\cite{imagenet_v2}. All models are trained and evaluated with the input resolution of $224\times224$.  Throughput is measured on one NVIDIA RTX 3090 GPU, with a batch size of 64 and averaged over 30 runs. Training time and testing time memory consumption is measured with a batch size of 64.}
\scalebox{0.7}{
\hskip-4em\begin{tabular}{l|ccc|cc|ccc}
Method & Params & FLOPs & Throughput (image/s) & Train Memory (MB) & Test Memory (MB) & ImageNet@Top-1 & Real@Top-1 & V2@Top-1 \\ \shline
ResNet-18  & 12M & 1.8G  & 4,454 & 2,812  & 1,511 & 69.8 & 77.3 & 57.1 \\
ResNet-50  & 26M & 4.1G  & 1,279 & 8,051  & 2,908 & 76.2 & 82.5 & 63.3 \\
ResNet-101 & 45M & 7.9G  & 722  & 10,710 & 3,053 & 77.4 & 83.7 & 65.7 \\ \hline
DeiT-Ti    & 5M  & 1.3G  & 3,398 & 2,170  & 314  & 72.2 & 80.1 & 60.4 \\
DeiT-S     & 22M & 4.6G  & 1,551 & 4,550  & 713  & 79.8 & 85.7 & 68.5 \\
DeiT-B     & 86M & 17.5G & 582  & 10,083 & 1,939 & 81.8 & 86.7 & 71.5 \\ \hline
PVT-Ti     & 13M & 1.9G  & 1,768 & 4,316  & 1,269 & 75.1 & 82.2 & 63.0 \\
PVT-S      & 25M & 3.8G  & 1,007 & 6,996  & 1,356 & 79.8 & 85.8 & 68.4 \\
PVT-M      & 44M & 6.7G  & 680  & 9,546  & 1,506 & 81.2 & 86.7 & 70.1 \\
PVT-L      & 61M & 9.8G  & 481  & 13,343 & 1,637 & 81.7 & 87.0 & 71.2 \\ \hline
Swin-Ti    & 28M & 4.5G  & 961  & 6,211  & 1,493 & 81.3 & 86.6 & 69.7 \\
Swin-S     & 50M & 8.7G  & 582  & 9,957  & 1,697 & 83.0 & 87.6 & 72.1 \\
Swin-B     & 88M & 15.4G & 386  & 13,705 & 2,453 & 83.3 & 87.7 & 72.3 \\ \hline
LIT-Ti     & 19M & 3.6G  & 1,294 & 5,868  & 1,194 & 81.1 & 86.6 & 70.4 \\
LIT-S      & 27M & 4.1G  & 1,298 & 5,973  & 1,264 & 81.5 & 86.4 & 70.4 \\
LIT-M      & 48M & 8.6G  & 638  & 12,248 & 1,473 & 83.0 & 87.3 & 72.0 \\
LIT-B      & 86M & 15.0G & 444  & 16,766 & 2,150 & 83.4 & 87.6 & 72.8
\end{tabular}
}
\label{tab:main_table}
\end{table*}

\begin{small}
	\begin{table*}[!ht]
		\begin{minipage}[t]{0.48\linewidth}
			\centering
            \renewcommand\arraystretch{1.2}
            \caption{Effect of our architecture design principle. * denotes the LIT model which adopts the same uniform patch merging strategies as in PVT-S or Swin-Ti.
            }
            \vspace{5pt}
            \scalebox{0.88}{
            \begin{tabular}{l|ccc}
            Model & Params & FLOPs & Top-1 Acc. (\%) \\ \shline
            PVT-S~\cite{pvt} & 25M & 3.8G & 79.8 \\
            \textbf{LIT-Ti*} & \bf{19M} & \textbf{3.6G} & \textbf{80.4} \\  \hline
            Swin-Ti~\cite{swin} & 28M & 4.5G & 81.3 \\
            \textbf{LIT-S*} & \bf{27M} & \textbf{4.1G} & 81.3 
            \end{tabular}
            }
            \label{tab:ablation_blk}
		\end{minipage}\hfill	
		\begin{minipage}[t]{0.48\linewidth}
			\centering
            \renewcommand\arraystretch{1.2}
            \caption{Effect of the proposed deformable token merging module. We replace the uniform patch merging strategies in PVT-S and Swin-Ti with our DTM.}
            \vspace{5pt}
            \scalebox{0.88}{
            \begin{tabular}{l|ccc}
            Model & Params & FLOPs & Top-1 Acc. (\%) \\ \shline
            PVT-S~\cite{pvt} & 25M & 3.8G & 79.8 \\
            \textbf{PVT-S + DTM} & 25M & 3.8G & \textbf{80.5} \\  \hline
            Swin-Ti~\cite{swin} & 28M & 4.5G & 81.3 \\
            \textbf{Swin-Ti + DTM} & 28M & 4.5G & \textbf{81.6}
            \end{tabular}
            }
            \label{tab:ablation_dtm}
		\end{minipage}
		\vspace{-10pt}
	\end{table*}
\end{small}

\paragraph{Results on ImageNet.}
In Table~\ref{tab:main_table}, we compare LIT with several state-of-the-art methods on ImageNet, ImageNet-Real and ImageNet-V2.
In general, all LIT models have fewer parameters, less FLOPs and faster throughput than their counterparts without applying any existing efficient self-attention mechanisms.  
For memory consumption, at the training time, we observe LIT-Ti and LIT-S require less memory than PVT-S and Swin-Ti while things are on the contrary when comparing LIT-M/B with Swin-S/B. The reason is due to the increased activation memory of standard MSA layers at the later stages. However, at the testing stage, LIT models show advantages over baselines as all of them consume less memory than their counterparts.

In terms of model performance, 
on ImageNet, LIT-Ti outperforms PVT-S by 1.3\% on the Top-1 accuracy while the FLOPs is reduced by 0.2G. LIT-S surpasses Swin-Ti by 0.2\% with 0.4G less FLOPs. LIT-M achieves on par performance with Swin-S, whereas the FLOPs of LIT-M is reduced. LIT-B brings 0.1\% Top-1 accuracy increase over Swin-B while using 0.4G less FLOPs. 
For ResNet and DeiT, LIT demonstrates better performance when compared to them with the same magnitude of FLOPs and parameters (\eg, ResNet-50 v.s. LIT-S, DeiT-B v.s. LIT-B).
On ImageNet-real, LIT-Ti improves PVT-S by 0.8\% on the Top-1 accuracy, while LIT-S/M/B achieve slightly lower accuracy than Swin models. Finally, on ImageNet-V2, LIT models achieve on par or better performance than PVT-S and Swin models. Overall, LIT models present competitive performance across the three datasets, challenging the full self-attention models in recent works.

\subsection{Ablation Studies}
\label{ablations}
\paragraph{Effect of the architecture design.}
To explore the effect of our architecture design principle in LIT, we conduct experiments on ImageNet and compare the architecture of LIT with two recent HVTs: PVT-S~\cite{pvt} and Swin-Ti~\cite{swin}. The results are shown in Table~\ref{tab:ablation_blk}.
In general, our architecture improves baseline PVT-S by 0.6\% in Top-1 accuracy while using less FLOPs (3.6G v.s. 3.8G). For Swin-Ti, our method reduces the FLOPs by 0.4G while achieving
on par performance. It is also worth noting that the total number of parameters is reduced for both PVT-S and Swin-Ti. The overall performance demonstrates the effectiveness of the proposed architecture, which also emphasizes the minor benefits of the early MSAs in PVT and Swin.

\begin{figure}[]
	\centering
	\includegraphics[width=0.8\linewidth]{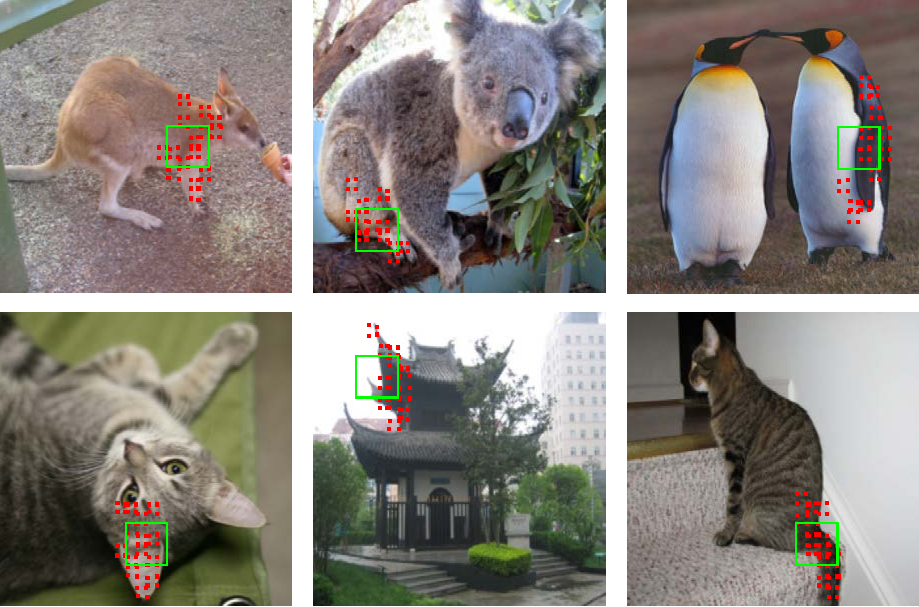}
	\caption{Visualization of learned offsets by the proposed deformable token merging modules. Each image shows $4^3$ sampling locations (\textit{red dots}) in three DTM modules with $2\times2$ filter. The green rectangle outlines a $32\times32$ patch of the original image, which also indicates a regular sampling field of previous methods. Best viewed in color. More examples can be found in the supplementary material.}
	\label{fig:dtm_vis}
\end{figure}

\paragraph{Effect of deformable token merging.}
To verify the effectiveness of our proposed DTM strategy, we replace the default patch merging scheme in PVT-S and Swin-Ti with DTM and train the models on ImageNet. The results are shown in Table~\ref{tab:ablation_dtm}. We observe that for both models, DTM introduces negligible FLOPs and parameters while improving PVT-S and Swin-Ti by 0.7\% and 0.3\% in terms of the Top-1 accuracy, respectively. Furthermore, we visualize the learned offsets in Figure~\ref{fig:dtm_vis}. As it shows, unlike previous uniform patch merging strategy where sampled locations are limited within the green rectangle, our DTM adaptively merges patches according to objects' scales and shapes (\eg, koala leg, cat tail). This again emphasizes the power of LIT to accommodate various geometric transformations.

\begin{figure*}[]
	\centering
	\includegraphics[width=0.9\linewidth]{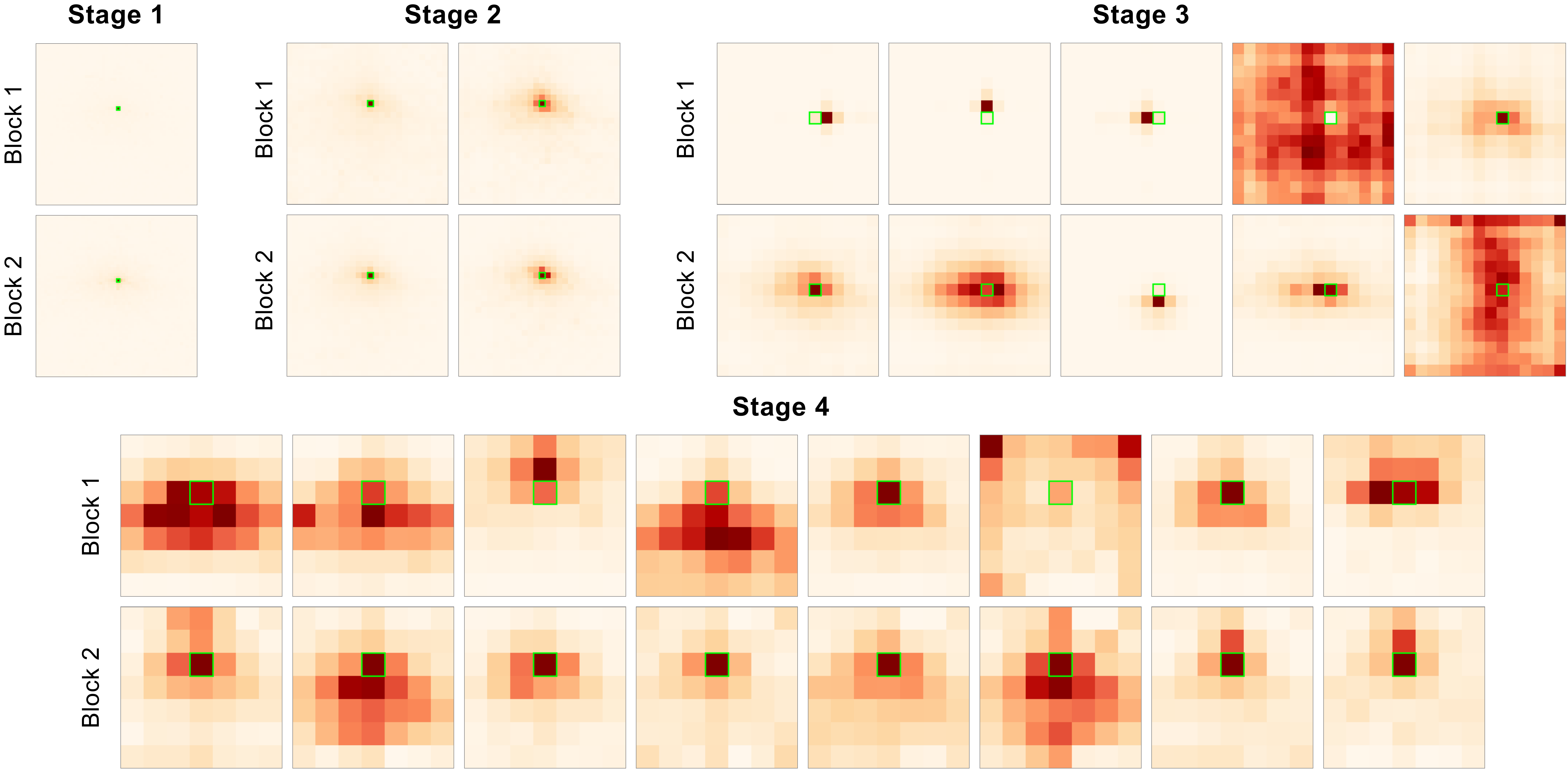}
	\caption{Attention probabilities of PVT-S with standard MSA layers. For each stage, we visualize the attention map of each head (\textit{columns}) at selected blocks (\textit{rows}).  All attention maps are averaged over 100 validation images. Each map shows the attention probabilities of a query pixel (\textit{green rectangle}) to other pixels. Darker color indicates higher attention probability and vice versa. Best viewed in color. We provide visualizations of all blocks in the supplementary material.}
	\label{fig:attn_prob_vis}
\end{figure*}

\paragraph{Effect of MSA in each stage.}
To explore the effect of self-attention in recent HVTs, we train PVT-S on ImageNet and gradually remove self-attention layers at each stage. The results are presented in Table~\ref{tab:ablation_stages}. First, after replacing the SRA layers in PVT-S with standard MSA layers, we observe 1.1\% improvement on the Top-1 accuracy whereas the FLOPs is almost doubled. This indicates that PVT makes a trade-off between performance and efficiency. Next, by gradually removing MSA layers in the first two stages, the Top-1 accuracy only drops by 0.1\%, 0.5\%, respectively. It implies that the self-attention layers in the early stages of PVT contribute less than expected to the final performance, and they perform not much better than pure MLP layers. It can be attributed to the fact that shallow layers focus more on encoding local patterns. However, we observe a huge performance drop when removing self-attention layers in the last two stages. The results show that the self-attention layers play an important role in the later stages and capturing long range dependencies is essential for well-performed hierarchical vision Transformers.

\begin{table}[]
\centering
\renewcommand\arraystretch{1.2}
\caption{Impact of the MSA layers at each stage. Note that PVT-S has four stages, which adopts SRA at all blocks instead of MSA. Here we denote ``w/ MSA'' as PVT-S with standard MSA layers. ``Stage'' refers to the stages where we remove all self-attention layers. For example, ``1,2'' means we remove the self-attention layers in the first two stages. Note that ``0'' means a model without removing any self-attention layers.}
\scalebox{0.8}
{
\begin{tabular}{l|cccc}
Model & Stage & Params & FLOPs& Top-1 Acc. (\%) \\ \shline
PVT-S~\cite{pvt} & 0 & 25M & 3.8G & 79.8 \\
PVT-S w/ MSA & 0 & 20M & 8.4G & 80.9 \\
PVT-S w/ MSA & 1 & 20M & 4.5G & 80.8 \\
PVT-S w/ MSA  & 1,2 & 19M & 3.6G & 80.4 \\
PVT-S w/ MSA  & 1,2,3 & 17M & 3.0G & 75.0 \\
PVT-S w/ MSA  & 1,2,3,4 & 14M & 2.8G & 66.8
\end{tabular}}
\label{tab:ablation_stages}
\vspace{-10pt}
\end{table}

To better understand the phenomenon, we visualize the attention probabilities for PVT-S without removing any MSA layers, which are depicted in Figure~\ref{fig:attn_prob_vis}. First, the attention map at the first stage shows that the query pixel almost pays no attention to other locations. At the second stage, the receptive field of the query pixel is slightly enlarged, but similar to the first stage. Considering that PVT-S only has one head at the first stage and two heads at the second stage, this strongly supports our hypothesis that very few heads in an MSA layer result in a smaller receptive field, such that a self-attention layer is almost equivalent to an FC layer. Furthermore, we observe relatively larger receptive fields from the attention maps of the last two stages. As a large receptive field usually helps to model longer dependencies, this explains the huge performance drop in Table~\ref{tab:ablation_stages} after we remove the MSA layers in the last two stages. 

\begin{table}[]
\centering
\renewcommand\arraystretch{1.2}
\caption{Object detection performance on the COCO \texttt{val2017} split using the RetinaNet framework.}
\scalebox{0.8}
	{
\begin{tabular}{l|ccccccc}
\multirow{2}{*}{Backbone} & \multicolumn{7}{c}{RetinaNet}                                                                     \\ \cline{2-8} 
                          & \multicolumn{1}{c|}{Params (M)} & $\mathrm{AP}$      & $\mathrm{AP}_{50}$   & \multicolumn{1}{c|}{$\mathrm{AP}_{75}$}   & $\mathrm{AP}_{S}$   & $\mathrm{AP}_{M}$   & $\mathrm{AP}_{L}$   \\ \shline
ResNet-50                 & \multicolumn{1}{c|}{38}         & 36.3    & 55.3 & \multicolumn{1}{c|}{38.6} & 19.3 & 40.0   & 48.8 \\
PVT-S                    & \multicolumn{1}{c|}{34}         & 40.4    & 61.3 & \multicolumn{1}{c|}{43.0} & 25.0 & 42.9 & 55.7 \\
LIT-Ti                    & \multicolumn{1}{c|}{\bf{30}}         & \bf{41.6} &  \bf{62.8}    & \multicolumn{1}{c|}{\bf{44.7}}     & \bf{25.7}     &  \bf{44.4}    &  \bf{56.4}   \\ \hline
ResNet-101                   & \multicolumn{1}{c|}{57}         & 38.5 & 57.8    & \multicolumn{1}{c|}{41.2}     & 21.4     &  42.6    &  51.1 \\
Swin-Ti                  & \multicolumn{1}{c|}{39}         & 41.5 & 62.1    & \multicolumn{1}{c|}{\bf{44.2}}     & 25.1     &  \bf{44.9}    &  55.5 \\
LIT-S               & \multicolumn{1}{c|}{39}       & \bf{41.6} & \bf{62.7}    & \multicolumn{1}{c|}{44.1}     &  \bf{25.6}    &  44.7    & \bf{56.5}  \\
\end{tabular}}
\label{tab:coco_retinanet}
\vspace{-5pt}
\end{table}

\begin{table}[]
\centering
\renewcommand\arraystretch{1.2}
\caption{Object detection and instance segmentation performance on the COCO \texttt{val2017} split using the Mask R-CNN framework. $\mathrm{AP}^b$ and $\mathrm{AP}^m$ denote the bounding box AP and mask AP, respectively.}
\scalebox{0.8}
{
\begin{tabular}{l|ccccccc}
\multirow{2}{*}{Backbone} & \multicolumn{7}{c}{Mask R-CNN}                                                                    \\ \cline{2-8} 
                          & \multicolumn{1}{c|}{Params (M)} & $\mathrm{AP}^{b}$       & $\mathrm{AP}^{b}_{50}$    & \multicolumn{1}{c|}{$\mathrm{AP}^{b}_{75}$}   & $\mathrm{AP}^{m}$      & $\mathrm{AP}^{m}_{50}$      & $\mathrm{AP}^{m}_{75}$   \\ \shline
ResNet-50                 & \multicolumn{1}{c|}{44}         & 38.0      & 58.6 & \multicolumn{1}{c|}{41.4} & 34.4 & 55.1 & 36.7 \\
PVT-S                   & \multicolumn{1}{c|}{44}         & 40.4    & 62.9 & \multicolumn{1}{c|}{43.8} & 37.8 & 60.1 & 40.3 \\
LIT-Ti                    & \multicolumn{1}{c|}{\bf{40}}         & \bf{42.0} & \bf{64.9}      & \multicolumn{1}{c|}{\bf{45.6}}     &   \bf{39.1}   & \bf{61.9}     &  \bf{41.9}  \\ \hline
ResNet-101                   & \multicolumn{1}{c|}{63}         & 40.4 & 61.1    & \multicolumn{1}{c|}{44.2}     & 36.4     &  57.7    &  38.8 \\
Swin-Ti                   & \multicolumn{1}{c|}{48}         & 42.2 & 64.6    & \multicolumn{1}{c|}{46.2}     & 39.1     &  61.6    &  42.0 \\
LIT-S               & \multicolumn{1}{c|}{48}       & \bf{42.9} &    \bf{65.6} & \multicolumn{1}{c|}{\bf{46.9}}     &   \bf{39.6}   &  \bf{62.3}    &  \bf{42.4} \\
\end{tabular}}
\label{tab:coco_maskrcnn}
\vspace{-10pt}
\end{table}

\subsection{Object Detection and Instance Segmentation on COCO}
In this section, we conduct experiments on COCO 2017~\cite{coco} dataset to show the performance of LIT on object detection and instance segmentation. COCO is a large-scale dataset which contains $\sim$118K images for the training set and $\sim$5K images for the validation set. For a fair comparison, we evaluate LIT-Ti and LIT-S on two base detectors: RetinaNet~\cite{retinanet} and Mask R-CNN~\cite{maskrcnn}.
For the experiments with both detectors, we consider 
CNN-based ResNet~\cite{resnet} and Transformer-based models including PVT-S~\cite{pvt} and Swin-Ti~\cite{swin}.
Following common practice~\cite{detr,pvt}, we measure the performance of all models by Average Precision (AP) in COCO.

\paragraph{Implementation details.}
All models are trained on 8 V100 GPUs, with $1\times$ schedule (12 epochs) and a total batch size of 16. We use AdamW~\cite{adamw} optimizer with a step decay learning rate scheduler. Following PVT~\cite{pvt}, the initial learning rates are set to $1\times10^{-4}$ and $2\times10^{-4}$ for RetinaNet and Mask R-CNN, respectively.
The weight decay is set to $1\times10^{-4}$ for all models. Results of Swin-Ti based detectors are adopted from Chu~\etal~\cite{chu2021Twins}. 
At the training stage, we initialize the backbone with the pretrained weights on ImageNet. The training images are resized to the shorter size of 800 pixels, and the longer size is at most 1333 pixels. During inference, we fix the shorter side of an image to 800 pixels.

\paragraph{Results on COCO.}
Table~\ref{tab:coco_retinanet} shows the comparisons of different backbones on object detection based on RetinaNet. 
By comparing LIT with PVT and ResNet counterparts, we find that our model outperforms both backbones on object detection in almost all metrics. Similar results can be found in Table~\ref{tab:coco_maskrcnn}, where LIT again surpasses compared methods on object detection and instance segmentation using the Mask R-CNN framework.

\subsection{Semantic Segmentation on ADE20K} \label{sem_seg}
We conduct experiments on ADE20K~\cite{ade20k} to show the performance of LIT models on semantic segmentation. ADE20K is a widely adopted dataset for semantic segmentation, which has $\sim$20K training images, $\sim$2K validation images and $\sim$3K test images. For a fair comparison, we evaluate LIT models with Semantic FPN~\cite{DKirillovGHD19}. Following the common practice in~\cite{pvt,swin}, we measure the model performance by mIoU.

\textbf{Implementation details.}
All models are trained on 8 V100 GPUs, with 8K steps and a total batch size of 16. The AdamW~\cite{adamw} optimizer is adopted with an initial learning rate of $1\times10^{-4}$. Learning rate is decayed by the polynomial decay schedule with the power of 0.9. We set the weight decay to $1\times10^{-4}$. All backbones are initialized with the pretrained weights on ImageNet. At the training stage, we randomly resize and crop the images to $512\times512$. During inference, images are scaled to the short size of 512.

\textbf{Results on ADE20K.}
We compare different backbones on the ADE20K validation set in Table~\ref{tab:ade20k}. From the results, we observe that LIT-Ti outperforms ResNet-50 and PVT-S by 4.6\% and 1.5\% mIoU, respectively. For LIT-S, our model again surpasses ResNet-101 and Swin-Ti, with 2.9\% and 0.2\% improvement on mIoU, respectively. The overall performance demonstrates the effectiveness of the proposed LIT models for dense prediction tasks.

\begin{table}[]
\centering
\renewcommand\arraystretch{1.2}
\caption{Semantic segmentation performance with different backbones on the ADE20K validation set.}
\vspace{-5pt}
\scalebox{0.8}
{
\begin{tabular}{l|cc}
\multirow{2}{*}{Backbone} & \multicolumn{2}{c}{Semantic FPN} \\ \cline{2-3} 
                          & Params (M)         & mIoU (\%)       \\ \shline
ResNet-50                & 29                 & 36.7        \\
PVT-S                     & 28                 & 39.8        \\
LIT-Ti                    & 24                 & \textbf{41.3}        \\ \hline
ResNet-101               & 48                 & 38.8        \\
Swin-Ti                  & 32                 & 41.5        \\
LIT-S                     & 32                 & \textbf{41.7}       
\end{tabular}
}
\vspace{-15pt}
\label{tab:ade20k}
\end{table}
\section{Conclusion and Future Work} \label{conclusion}
In this paper, we have introduced LIT, a hierarchical vision transformer which pays less attention in the early stages to ease the huge computational cost of self-attention modules over high-resolution representations. Specifically, 
LIT applies MLP blocks in the first two stages to focus on local patterns while employing standard Transformer blocks with sufficient heads in the later stages to handle long range dependencies.
Moreover, we have proposed a deformable token merging module, which is learned to adaptively merge informative patches to an output unit, with enhanced geometric transformations. 
Extensive experiments on ImageNet, COCO and ADE20K have demonstrated that LIT achieves better performance
compared with existing state-of-the-art HVT methods.
Future works may include finding better architectural configurations of LIT with neural architecture search (NAS) and improving MLP blocks in the early stages to enhance the capability of LIT to encode local patterns.
Besides, one may also consider applying efficient self-attention mechanisms in the later stages to achieve better efficiency, such as kernelization~\cite{peng2021random}, low-rank decomposition \cite{linformer}, memory \cite{rae2020compressive}, and sparsity \cite{sparse_attn} schemes, etc. 
\section{Appendix}

\subsection{Architecture Details} \label{sec:arch}
 Table~\ref{tab:model_arch} shows the architecture specifications of LIT. In general, LIT-Ti and LIT-S/M/B have the same depth, width and MLP expansion ratio configurations as PVT-S and Swin-Ti/S/B, respectively. Besides, for a fair comparison, the kernel sizes of DTM in LIT models are consistent with those of patch merging layers in their counterparts.
 
 \begin{table*}[]
\centering
\renewcommand\arraystretch{1.2}
\caption{Architecture specifications of LIT.  $P$ denotes the patch size and $C$ is the channel dimension. $L$ refers to the number of blocks applied at each stage. $N$ is the number of self-attention heads. As the first two stages do not have self-attention layers, we only show the numbers of $N_3$ and $N_4$. The expansion ratios of LIT-Ti at each stage are [8, 8, 4, 4]. We use an expansion ratio of 4 for all MLP layers in LIT-S, LIT-M and LIT-B. The numbers denoted under each stage index (\eg, $28 \times 28$) represents the size of the feature maps at that stage, based on the input resolution of $224 \times 224$.}
\scalebox{1.0}
{
\begin{tabular}{c|c|l|l|l|l}
 Stage &
  Layer Name &
  \multicolumn{1}{c|}{LIT-Ti} &
  \multicolumn{1}{c|}{LIT-S} &
  \multicolumn{1}{c|}{LIT-M} &
  \multicolumn{1}{c}{LIT-B} \\ \shline
\multirow{2}{*}{\begin{tabular}[c]{@{}c@{}}Stage 1\\ $56\times56$\end{tabular}} &
  Patch Embedding &
  $\begin{array}{l}P_1=4\\C_1 = 64\end{array}$ &
  $\begin{array}{l}P_1=4\\C_1 = 96\end{array}$ &
  $\begin{array}{l}P_1=4\\C_1 = 96\end{array}$ &
  $\begin{array}{l}P_1=4\\C_1 = 128\end{array}$ \\ \cline{2-6} 
 &
  MLP Block &
  $\begin{array}{l}L_1 = 3\end{array}$ &
  $\begin{array}{l}L_1 = 2\end{array}$ &
  $\begin{array}{l}L_1 = 2\end{array}$ &
  $\begin{array}{l}L_1 = 2\end{array}$ \\ \hline
\multirow{2}{*}{\begin{tabular}[c]{@{}c@{}}Stage 2\\ $28\times28$\end{tabular}} &
  DTM &
  $\begin{array}{l}P_2=2\\C_2 = 128\end{array}$ &
  $\begin{array}{l}P_2=2\\C_2 = 192\end{array}$ &
  $\begin{array}{l}P_2=2\\C_2 = 192\end{array}$ &
  $\begin{array}{l}P_2=2\\C_2 = 256\end{array}$ \\ \cline{2-6} 
 &
  MLP Block &
  $\begin{array}{l}L_2 = 4\end{array}$ &
  $\begin{array}{l}L_2 = 2\end{array}$ &
  $\begin{array}{l}L_2 = 2\end{array}$ &
  $\begin{array}{l}L_2 = 2\end{array}$ \\ \hline
\multirow{3}{*}{\begin{tabular}[c]{@{}c@{}}Stage 3\\ $14\times14$\end{tabular}} &
  DTM &
  $\begin{array}{l}P_3=2\\C_3 = 320\end{array}$ &
  $\begin{array}{l}P_3=2\\C_3 = 384\end{array}$ &
  $\begin{array}{l}P_3=2\\C_3 = 384\end{array}$ &
  $\begin{array}{l}P_3=2\\C_3 = 512\end{array}$ \\ \cline{2-6} 
 &
  Transformer Block &
  $\begin{array}{l}N_3 = 5\\L_3 = 6\end{array}$ &
  $\begin{array}{l}N_3 = 12\\L_3 = 6\end{array}$ &
  $\begin{array}{l}N_3 = 12\\L_3 = 18\end{array}$ &
  $\begin{array}{l}N_3 = 16\\L_3 = 18\end{array}$ \\ \hline
\multirow{3}{*}{\begin{tabular}[c]{@{}c@{}}Stage 4\\ $7\times7$\end{tabular}} &
  DTM &
  $\begin{array}{l}P_4=2\\C_4 = 512\end{array}$ &
  $\begin{array}{l}P_4=2\\C_4 = 768\end{array}$ &
  $\begin{array}{l}P_4=2\\ C_4 = 768\end{array}$ &
  $\begin{array}{l}P_4=2\\C_4 = 1024\end{array}$ \\ \cline{2-6} 
 &
  Transformer Block &
  $\begin{array}{l}N_4 = 8\\L_4 = 3\end{array}$ &
  $\begin{array}{l}N_4 = 24\\L_4 = 2\end{array}$ &
  $\begin{array}{l}N_4 = 24\\L_4 = 2\end{array}$ &
  $\begin{array}{l}N_4 = 32\\L_4 = 2\end{array}$ \\
\end{tabular}
}
\label{tab:model_arch}
\end{table*}
 
 \subsection{Visualizations} \label{sec:visualization}
\subsubsection{Visualizations of Attention Probabilities}
\begin{figure*}[]
	\centering
	\includegraphics[width=\linewidth]{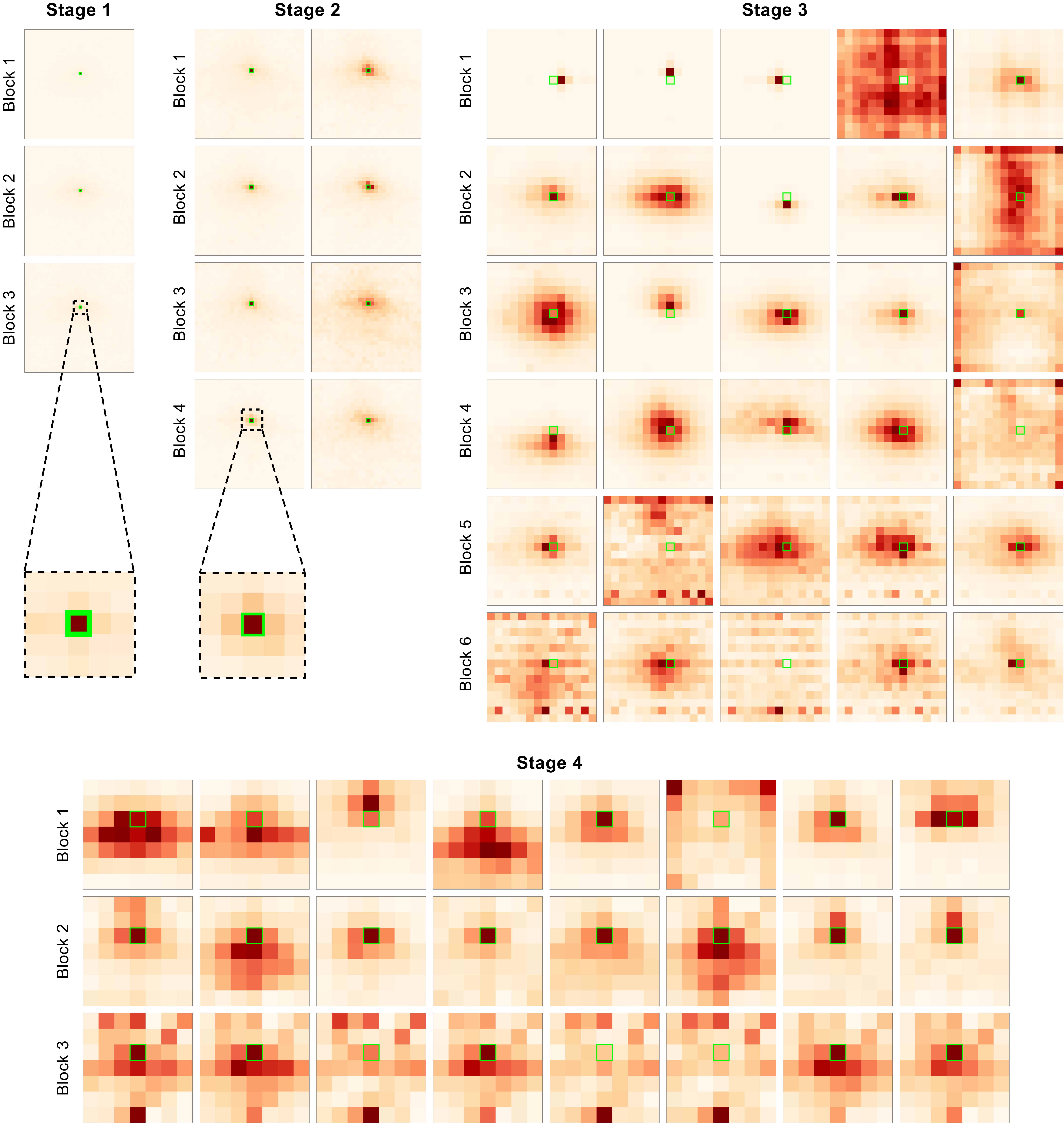}
	\caption{Attention probabilities of PVT-S with standard MSA in all Transformer blocks. Best viewed in color.}
	\label{fig:attn_probs_all}
\end{figure*}

In Figure~\ref{fig:attn_probs_all}, we visualize the attention probabilities of MSA layers in all Transformer blocks from our modified PVT-S with standard MSA layers. As it shows, early MSA layers only attend to a tiny area of around the query pixel, especially for the first two stages. We also observe that the MSA layers in the later two stages have a quite large attention area, in which case we keep the later MSA layers in LIT for better performance.

\subsubsection{Visualizations of Learned Offsets}
\begin{figure*}[]
	\centering
	\includegraphics[width=\linewidth]{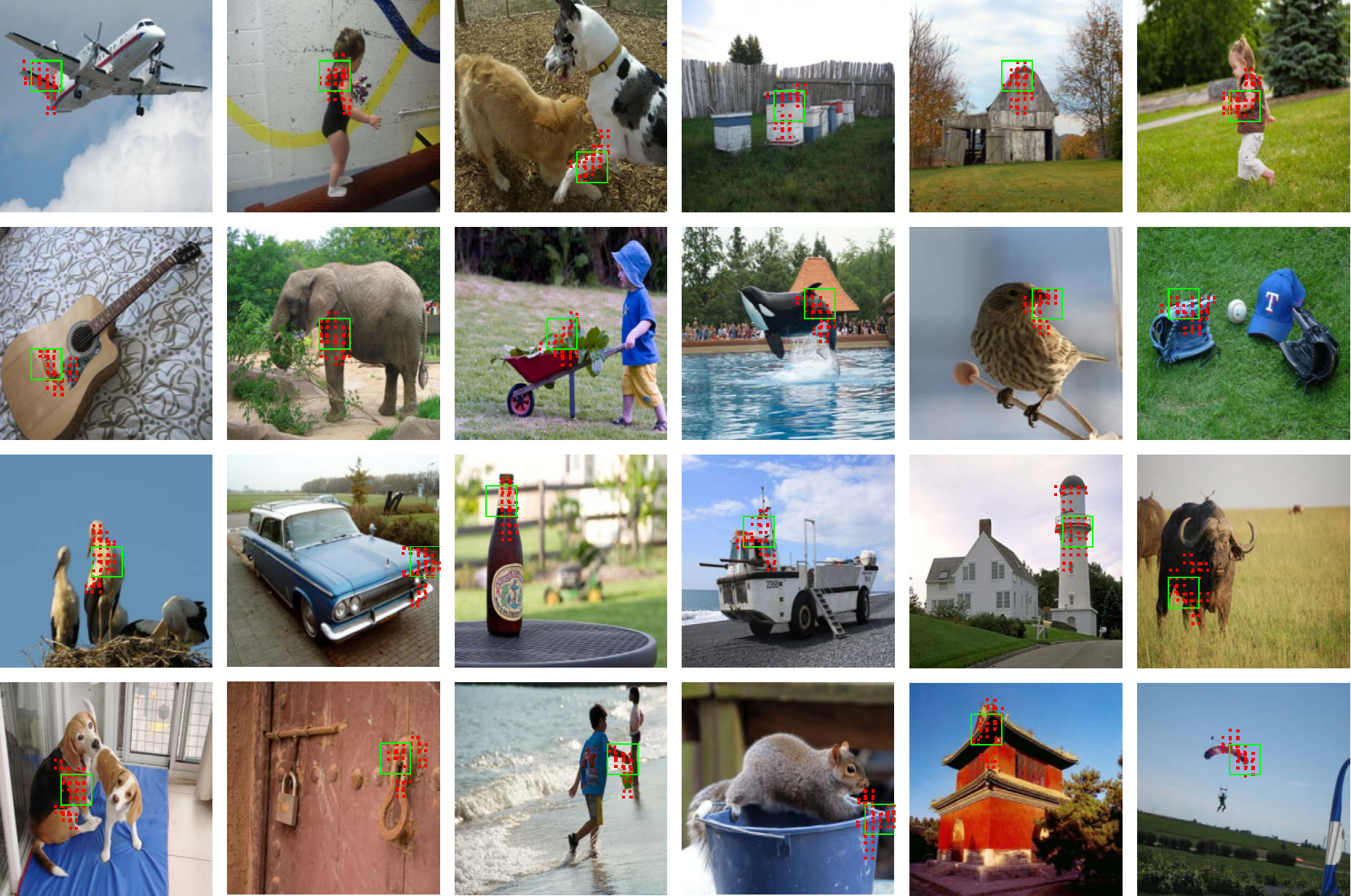}
	\caption{More examples of learned offsets by the proposed deformable token merging modules.}
	\label{fig:dtm_vis_full}
\end{figure*}

In Figure~\ref{fig:dtm_vis_full}, we provide more examples of the learned offsets by our proposed DTM. We can observe from the visualizations that DTM can adapt to a variety of objects' shapes and transformations.

\newpage
\bibliographystyle{ieee_fullname.bst}
\bibliography{citation}

\end{document}